\documentclass[12pt]{article}

\usepackage[a4paper, margin=1in]{geometry}
\usepackage{setspace}
\usepackage[utf8]{inputenc}
\usepackage[T1]{fontenc}

\usepackage{amsmath}
\usepackage{amsfonts}
\usepackage{nicefrac}
\usepackage{microtype}
\usepackage{xcolor}

\usepackage{graphicx}
\usepackage{subfig}
\usepackage{booktabs}
\usepackage{float}

\usepackage{enumitem}
\usepackage{url}
\usepackage{natbib}
\usepackage{authblk}

\usepackage{hyperref}

\title{\textbf{CTR Prediction on Alibaba's Taobao Advertising Dataset Using Traditional and Deep Learning Models}}

\author{
\begin{minipage}{0.25\textwidth}
\centering
Hongyu Yang \\

\end{minipage}
\hfill
\begin{minipage}{0.25\textwidth}
\centering
Chunxi Wen \\

\end{minipage}
\hfill
\begin{minipage}{0.25\textwidth}
\centering
Jiyin Zhang \\

\end{minipage}
\\
\vspace{0.5cm}
\begin{minipage}{0.25\textwidth}
\centering
Nanfei Shen \\

\end{minipage}
\hfill
\begin{minipage}{0.25\textwidth}
\centering
Shijiao Zhang \\

\end{minipage}
\hfill
\begin{minipage}{0.25\textwidth}
\centering
Xiyan Han \\

\end{minipage}
\vspace{0.5cm}

Department of Biostatistics, Data Science in Health \\
University of California, Los Angeles
}

\date{}

\begin{document}

\maketitle

\begin{abstract}
Click-through rates prediction is critical in modern advertising systems, where ranking relevance and user engagement directly impact platform efficiency and business value. In this project, we explore how to model CTR more effectively using a large-scale Taobao dataset released by Alibaba.

We start with supervised learning models, including logistic regression and LightGBM, that are trained on static features such as user demographics, ad attributes, and contextual metadata. These models provide fast, interpretable benchmarks, but have limited capabilities to capture patterns of behavior that drive clicks.

To better model user intent, we combined behavioral data from hundreds of millions of interactions over a 22-day period. By extracting and encoding user action sequences, we construct representations of user interests over time. We use deep learning models to fuse behavioral embeddings with static features. Among them, multilayer perceptrons (MLPs) have achieved significant performance improvements.

To capture temporal dynamics, we designed a Transformer-based architecture that uses a self-attention mechanism to learn contextual dependencies across behavioral sequences, modeling not only what the user interacts with, but also the timing and frequency of interactions. Transformer improves AUC by 2.81 \% over the baseline (LR model), with the largest gains observed for users whose interests are diverse or change over time.

In addition to modeling, we propose an A/B testing strategy for real-world evaluation. We also think about the broader implications: personalized ad targeting technology can be applied to public health scenarios to achieve precise delivery of health information or behavior guidance. Our research provides a roadmap for advancing click-through rate predictions and extending their value beyond e-commerce.

\end{abstract}
\newpage
\section{Introduction and Problem Statement}

As online advertising systems have gradually evolved from traditional broadcast-style delivery to programmatic trading models, click-through rate (CTR) prediction has become a core task for ad ranking and delivery optimization. In a typical cost-per-click (CPC) mechanism, the display order of ads is usually determined by the product of the estimated click-through rate (CTR) and the bid, in order to maximize the expected revenue. The precision of CTR prediction directly impacts ad ranking, bidding strategies, platform revenue, and user experience (\cite{yang2024aie}). Early CTR prediction models primarily relied on linear models such as Logistic Regression (LR) due to computational efficiency and ease of deployment, it has become the industry benchmark model. However, LR models have limitations in handling high-dimensional sparse data and complex nonlinear feature interactions, making it difficult to capture the deep relationships between users, ads, and context. To overcome these challenges, researchers have proposed various deep learning models, for example, Alibaba further proposed the Deep Interest Network (DIN), which incorporates an attention mechanism to dynamically capture user interests, achieving a significant 9.2\% improvement in AUC within the Taobao advertising system (\cite{zhou2018deep}).

This project was conducted in collaboration between the UCLA team and the Microsoft Ads Team, aiming to explore more practical CTR modeling approaches in real-world advertising systems. The primary objective is to build and evaluate models using the publicly available CTR dataset provided by the Alibaba Tianchi platform, build efficient and scalable CTR prediction models by integrating user features, ad features, and contextual information, and applying various machine learning methods to model the probability of ad clicks, the efficiency of ad matching can be significantly improved.

\section{Data Collection and Pre-Processing}

\subsection{Data Sources and Key Attributes}

The dataset used in this project was released by Alibaba Group via the Tianchi platform, covering user–ad interactions on Taobao from April 16 to May 13, 2017. It consists of four main structured tables:

\textbf{raw\_sample.csv}: Each row represents a single ad impression, with key fields including \texttt{user\_id}, \texttt{adgroup\_id}, \texttt{clk} (click label), and \texttt{time\_stamp}. There are approximately 26.6 million records.

\textbf{ad\_feature.csv}: Contains 846,811 rows of static ad features, such as product category, brand, and price.

\textbf{user\_profile.csv}: Includes nine user characteristics (e.g., gender, age level, consumption ability, shopping depth, city tier). Each row corresponds to an ad impression, and is indexed by composite keys including \texttt{user\_id}, \texttt{adgroup\_id}, and \texttt{time\_stamp}.

\textbf{user\_behavior\_log.csv}: Logs user behaviors (click, cart, favorite, purchase) from April 16 to May 5, 2017, totaling about 704 million records.

\subsection{Data Quality Issues and Preprocessing Strategy}

The overall click-through rate (CTR) is only 5.14\%, indicating a significant class imbalance. We applied sampling strategies to mitigate this and improve model performance.

Some fields have missing values. For example, \texttt{pvalue\_level} has the highest missing rate (54.8\%), followed by \texttt{brand} (31.2\%) and \texttt{new\_user\_class\_level} (31.0\%). Profile fields such as \texttt{gender}, \texttt{age\_level}, and \texttt{shopping\_level} have about 5.8\% missingness, primarily due to data source limitations.

Missing values were handled based on model type: in logistic regression, categorical variables were filled with "-1" and price with the median; in LightGBM, categorical variables were filled with "missing" and price also with the median. For MLP, categorical features were filled with the mode and numeric features (like price) with the median to better preserve distribution.

\section{Methods}
\subsection{Exploratory Data Analysis (EDA)}

\subsubsection{User Feature Analysis}

We analyzed how user characteristics such as age, gender, spending power, and location relate to CTR and shopping behavior. The overall CTR is 5.14\%, with female users clicking slightly more (5.24\%) than males (4.83\%). CTR remains stable across weekdays and weekends, and also across time of day, despite peak user activity occurring around 10pm.

\begin{figure}[htbp]
\centering
\subfloat[City Tier Distribution]{
    \includegraphics[width=0.31\linewidth]{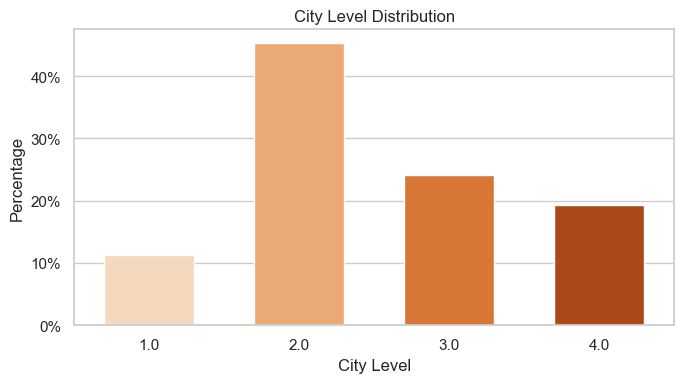}
}
\hfill
\subfloat[Consumption by City]{
    \includegraphics[width=0.31\linewidth]{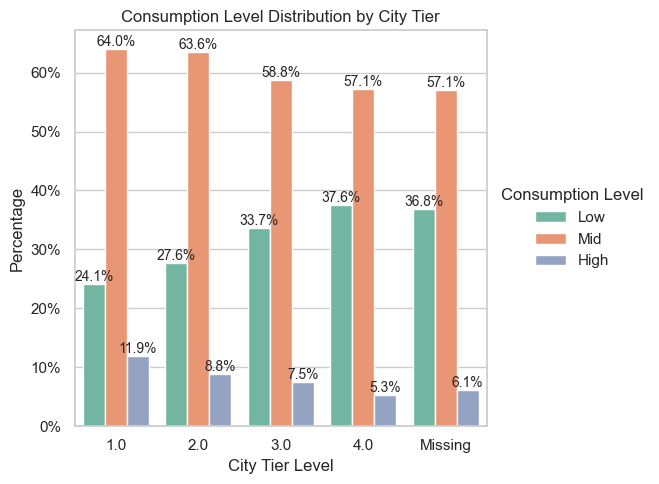}
}
\hfill
\subfloat[Shopping Depth by City]{
    \includegraphics[width=0.31\linewidth]{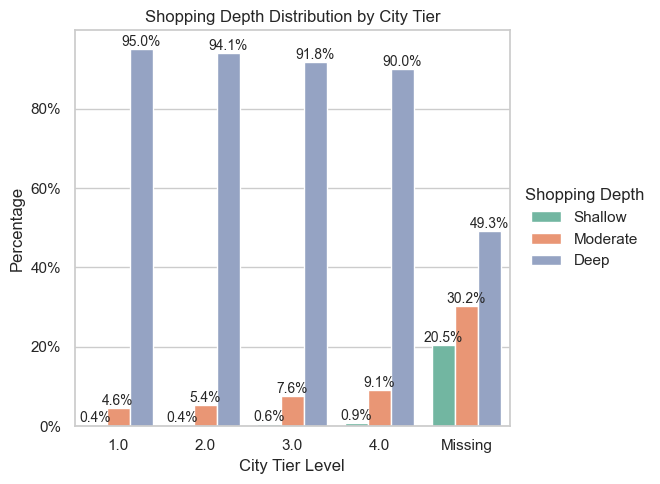}
}
\caption{User behavior across city tiers}
\label{fig:city_profile}
\end{figure}

\paragraph{City Tier Analysis}
As shown in Figure~\ref{fig:city_profile}, users from tier-2 cities make up the largest segment. Interestingly, users in higher-tier cities tend to have lower high-consumption percentages and higher low-consumption proportions. Deep shopping behavior, however, is prevalent across all city tiers.

\begin{figure}[htbp]
\centering
\subfloat[Consumption by Gender]{
    \includegraphics[width=0.45\linewidth]{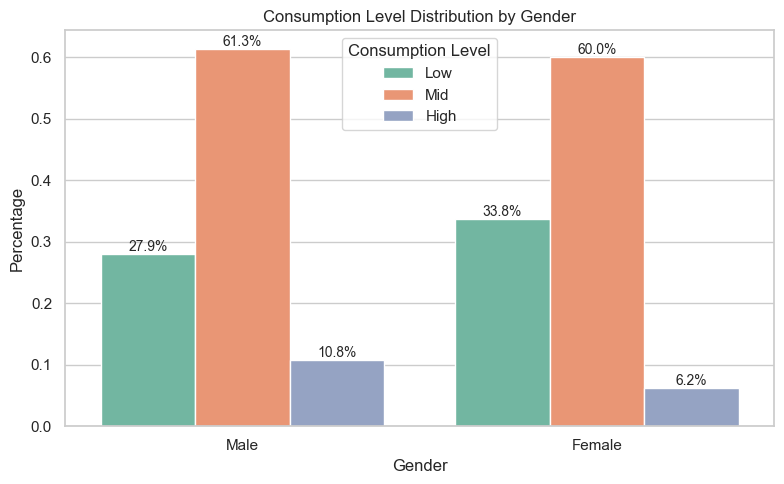}
}
\hfill
\subfloat[Shopping Depth by Gender]{
    \includegraphics[width=0.45\linewidth]{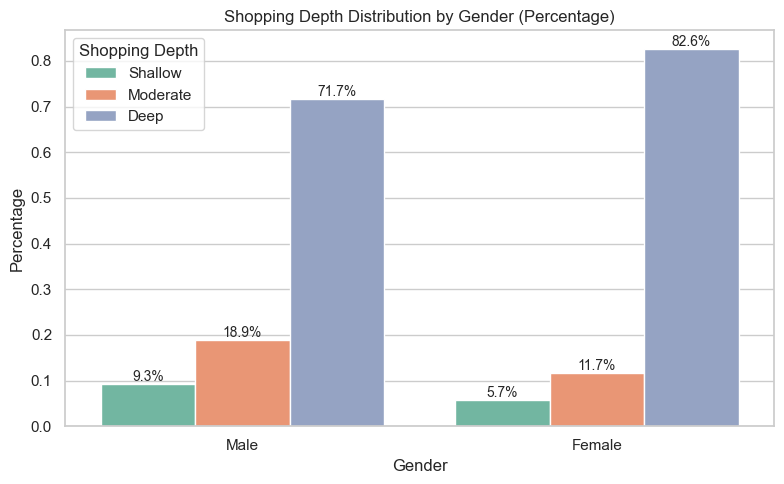}
}
\caption{Consumption level and shopping depth by gender}
\label{fig:gender_profile}
\end{figure}

\paragraph{Gender-Based Differences}
Figure~\ref{fig:gender_profile} shows that male and female users exhibit similar spending and shopping depth patterns. Mid-level consumption dominates both groups (~60\%). Female users spend more on low-level consumption (+5.9\%), while males spend slightly more at the high tier (+4.6\%). Deep shoppers dominate across both genders.

\paragraph{Feature Engineering}
To incorporate these patterns into our model, we derived variables such as \texttt{is\_weekend} and \texttt{time\_bin}. All categorical variables are label encoded and passed through an embedding layer. For high-cardinality variables like \texttt{age\_level}, infrequent values are grouped as “other” to reduce sparsity. Continuous features such as click counts are normalized.

\subsubsection{Ad Feature Analysis}

The advertising dataset consists of six variables: ad group ID, category, campaign, customer, brand, and price. Among the 846,800 ad groups, the brand variable contains 29.09\% missing values, which we impute using the mode.

From exploratory analysis, several features exhibit strong correlation with click-through rate (CTR). Notably, the brand-level CTR distribution shows clear heterogeneity. Some heavily advertised brands have lower CTRs than less exposed ones, suggesting that advertising effectiveness is influenced not only by brand popularity but also by placement strategies and user-ad matching.

We also observed a steady decline in CTR as price increases. To capture this trend, we bucket prices into seven levels (Very Low to Super Luxury), which were later used as derived features.

Finally, we visualize the CTR distribution among brands with over 1,000 exposures, showing its long-tailed, right-skewed nature. This pattern highlights both variability in brand impact and modeling opportunities, such as identifying outlier brands or engineering CTR-based features.

\subsubsection{Contextual Feature Analysis}

\begin{figure}[htbp]
\centering

\subfloat[Ad Clicks by Day (May 6–13)]{
    \includegraphics[width=0.48\linewidth]{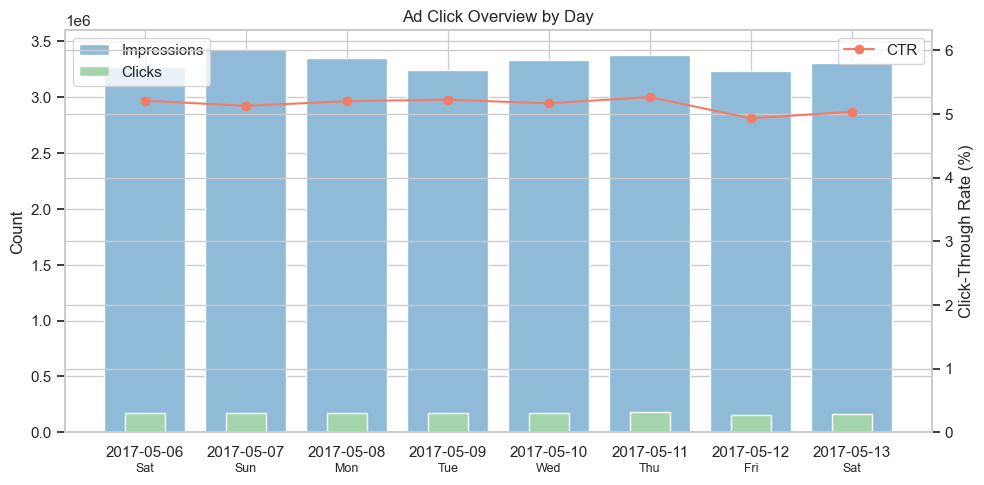}
    \label{fig:sub_day}
}
\hfill
\subfloat[Ad Clicks by Hour of Day]{
    \includegraphics[width=0.48\linewidth]{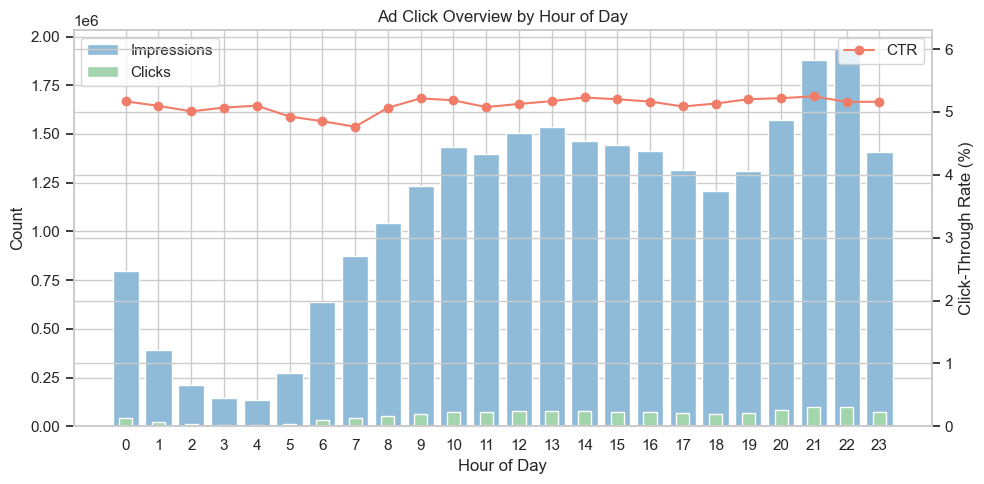}
    \label{fig:sub_hour}
}

\caption{Temporal patterns in impressions, clicks, and CTR}
\label{fig:time}
\end{figure}

We examined time-related features by day and hour to understand user engagement patterns. The original timestamps were converted from UTC seconds to Beijing time for interpretability. As shown in Figure~\ref{fig:time}, impressions and clicks were stable from May 6 to May 13, with CTR consistently around 5\% and a slight dip on Friday. Across hours of the day, user activity rises from 6 am and peaks around 10 pm, but CTR remains stable, indicating balanced ad performance throughout the day.

\subsubsection{Behavior Feature Analysis}

To enrich our understanding of user interaction patterns beyond static profiles, we conducted exploratory analysis on the \texttt{user\_behavior\_log.csv} dataset, which contains over 700 million behavior records collected from April 16 to May 5, 2017. Each record logs a user’s action — such as page view (pv), add to cart (cart), favorite (fav), or purchase (buy) — along with associated category and brand information. As expected, browsing behavior (\texttt{pv}) dominates the dataset, while high-intent actions like \texttt{cart} and \texttt{buy} occur much less frequently, consistent with the typical conversion funnel observed in e-commerce platforms.

In terms of temporal trends, user activity shows stable daily volume with mild fluctuations, and distinct hourly patterns: activity ramps up around 6 a.m., peaks between 9–10 p.m., and drops off late at night. Interestingly, although behavior frequency varies over time, the proportion of higher-intent actions remains relatively steady, suggesting that while users are more active in the evening, their likelihood of converting does not change significantly by hour.

User behavior also varies substantially across individuals. While many users only have a few recorded actions, a small group contributes disproportionately high volumes of interactions. This long-tailed distribution indicates the presence of both light and heavy users, and highlights the need for appropriate sequence processing techniques to avoid biasing the model toward hyperactive users.

Finally, we examined category-level behavior patterns and found that some product categories — such as electronics and cosmetics — exhibit stronger downstream signals, with higher ratios of \texttt{cart} and \texttt{buy} behaviors. These findings provide useful signals for interest modeling and support the integration of behavior sequences into downstream CTR prediction models using sequence-aware architectures.

\subsection{Static Feature-Based Models}
Our modeling pipeline includes multiple stages, progressing from simple linear models to advanced deep learning methods. We first implemented three classical models: Logistic Regression, LightGBM, and a neural network with a multilayer perceptron (MLP) architecture. These serve as strong baselines commonly used in CTR prediction tasks (\cite{yang2022click}). Later, based on analysis of user behavior logs, we extended our modeling to include Transformer-based architectures to better capture sequential behavioral patterns.

\subsubsection{Logistic Regression Model}

 In feature engineering, we eliminated the overly dimensional categorical variables \texttt{user\_id} (1.14 million unique values) and \texttt{adgroup\_id} (840,000 unique values) and introduced two CTR-based statistical traits: \texttt{ad\_ctr\_cv} (click-through rate) and \texttt{user\_ctr\_cv} (click-through rate) to reflect the historical behavioral tendencies of ads and users. These features are very common in CTR estimation and can improve model performance. Unlike using IDs directly, CTR features provide a more continuous and interpretable signal that helps the model identify which samples are more likely to be clicked. In order to avoid data leakage, we used a cross-validated encoding strategy when constructing these two features, that is, only the CTR was calculated on the training set and then applied to the validation set, supplemented by smoothing to reduce the unstable effect caused by too few exposures. In addition, we extracted and transformed the \texttt{time\_stamp} variables to generate \texttt{hour} and \texttt{weekday}.

In the end, we included categorical variables including: \texttt{final\_gender\_code}, \texttt{age\_level}, \texttt{occupation}, \texttt{shopping\_level}, \texttt{cate\_id\_filtered}, \texttt{pid}, \texttt{hour}, \texttt{weekday}.

Numeric variables include: \texttt{ad\_ctr\_cv}, \texttt{user\_ctr\_cv} and \texttt{price}.

To address the problem of a severely unbalanced sample (click-through rate of only about 5\%), instead of using the "balanced" parameter, we manually calculated the proportion of positive and negative samples and set the weight of positive examples. At the same time, we set the regularization intensity to \texttt{C=0.1} to reduce the risk of overfitting. After the model was fitted to the training set, the performance was evaluated on the validation set.

The final results show that the model AUC is 0.659 and PR AUC is 0.097 at the default threshold, and the F1 score is improved to 0.157, Precision is 0.105, Recall is 0.313, and the accuracy is 0.828 at the optimal threshold (about 0.593).

\subsubsection{LightGBM Model}

We continue feature processing in Logistic Regression and introduce two statistical features of the CTR: 
\texttt{ad\_ctr\_cv} (click-through rate of ads) and 
\texttt{user\_ctr\_cv} (click-through rate of users), 
as well as the method of extracting time variables from 
\texttt{time\_stamp}. 

In addition, we've included a number of user-to-product interaction characteristics, including: 

\texttt{final\_gender\_code\_cate\_id\_filtered\_inter},

\texttt{age\_level\_cate\_id\_filtered\_inter},

\texttt{brand\_cate\_id\_filtered\_inter}. 

To avoid sparsity issues, we keep only the first few of the most common combinations of each class (such as the top 100 or 150), and the rest are classified as ``other''.

The categorical variables we ultimately included were: \texttt{final\_gender\_code}, \texttt{age\_level}, \texttt{occupation}, \texttt{shopping\_level}, \texttt{cate\_id\_filtered}, \texttt{brand}, \texttt{campaign\_id}, \texttt{hour}, \texttt{weekday}, \texttt{cms\_segid}, 
\texttt{cms\_group\_id}, \texttt{pvalue\_level}, \texttt{new\_user\_class\_level}, 

\texttt{final\_gender\_code\_cate\_id\_filtered\_inter}, \texttt{age\_level\_cate\_id\_filtered\_inter}, \texttt{brand\_cate\_id\_filtered\_inter};

Numeric variables include: \texttt{ad\_ctr\_cv}, \texttt{user\_ctr\_cv} and \texttt{price}.

We identified the optimal combination of parameters for LightGBM through grid search: learning\_rate=0.1, max\_depth=5, num\_leaves=31, and reg\_lambda=100, and built the final model base on this and used the full training data for training. In the validation set evaluation, we applied an early stopping strategy and used AUC and LogLoss as evaluation metrics.

The final model achieved the following performance on the validation set: AUC = 0.6598, PR-AUC = 0.0991, Log Loss = 0.1945, F1 score of 0.1602, accuracy of 0.8384, precision and recall of 0.1093 and 0.2995, respectively. The feature importance analysis shows that \texttt{user\_ctr\_cv}, \texttt{campaign\_id}, and brand are among the most critical features, and additionally, the three interaction features we constructed are also ranked within top 20, indicating that they have a positive effect on the performance of the model.

\subsubsection{Neural Network Model}

In order to further explore the nonlinear relationship between user behavior and advertising characteristics, we constructed and trained a PyTorch-based multi-layer perceptron (MLP) model in this project to predict the probability of users clicking on advertisements. This model, with the structure of embedding + dense layer, is capable of learning deep-level nonlinear mapping relationships from sparse and high-dimensional inputs.

Considering the huge number of categories of advertising-related features (such as brand, campaign\_id, customer, etc.), if One-Hot encoding is directly carried out, it will cause a dimension explosion and the computing and storage costs will be too high. Therefore, we first uniformly perform Label Encoding on all categoric variables and send them to the Embedding layer to map each category to a low-dimensional dense vector. It can be understood that the model learns a "compact fingerprint" for each category and compresses to represent its semantic information.

In terms of numerical variables (such as price, hour, etc.), we use standardization (Z-score) processing to ensure that variables of different dimensions are input into the model at the same scale. Meanwhile, we uniformly encode the values with extremely low occurrence frequency in all category features as "\_\_other\_\_" to reduce long-tail interference and improve generalization ability.

The overall structure of the model is as follows: After all the vectors output by the embedding layer are concatenated with numerical features, they enter a three-layer fully connected network (containing 128, 64 and 1 neuron respectively). After each layer, Batch Normalization and Dropout are followed to enhance the training stability and prevent overfitting. The output layer is a Sigmoid unit, which is used to output the final click probability.

During the training process, we adopt Binary Cross-Entropy as the loss function and use the Adam optimizer for parameter update. The training data is divided into the training set and the validation set in an 8:2 ratio. The batch size is 1024, and a total of 10 epochs are trained. The training process is carried out on the GPU, and the total time consumption is approximately 90 minutes.

On the premise of not introducing historical CTR features in the test set, the MLP model achieved excellent performance on the validation set: the AUC was 0.663, the Log Loss was 0.638, and the PR AUC reached 0.101,[Table \ref{models-table}] which was significantly better than the previous two models. The results show that the MLP model can not only capture complex feature combination relationships, but also has a strong generalization ability, especially excelling in the modeling of category variables and the mining of user behavior patterns.

\subsection{Behavior-Augmented Models with Sequence Modeling}

Following the evaluation of baseline models with static features, we developed enhanced architectures that incorporate user behavior sequences to capture temporal interaction patterns and improve prediction accuracy.

\subsubsection{Deep Neural Network with Behavior Sequences}

Our first behavior-augmented model extends the traditional feature-based MLP approach by integrating user historical behavior sequences. This extension addresses the limitation of static feature representations by capturing sequential patterns in user interactions.

\textbf{Architecture Overview}

The model processes three types of information: traditional categorical and numerical features from our baseline models, plus a new component that captures user behavior sequences. Each user's historical interactions are represented as sequences of (category, brand) pairs, creating a temporal storyline of their preferences and browsing patterns.

\textbf{Behavior Sequence Processing}

User behavior sequences present unique challenges in CTR prediction. Different users have vastly different interaction histories - some might have hundreds of recorded behaviors while others have only a few. To address this, we standardize all sequences to length 20, padding shorter sequences and keeping only the most recent 20 interactions for longer ones.

Each interaction contains both category information (electronics, clothing, books) and brand information (Nike, Apple, Samsung). We create separate embedding spaces for categories and brands, then combine them to capture the joint preference patterns. The intuition is that a user interested in "electronics + Apple" might have different click behavior than someone interested in "electronics + Samsung."

Rather than complex aggregation schemes, we simply average the sequence representations to create a single user behavior vector. This approach proves surprisingly effective while maintaining computational efficiency.

\textbf{Network Design}

The final architecture brings together three streams of information. Categorical features like user demographics and ad properties go through embedding layers to create dense representations. Numerical features like historical CTR and time-based patterns get linearly transformed to match the embedding dimensions. The behavior sequence vector from our temporal modeling joins these components.

All representations are concatenated and fed through a deep neural network with batch normalization and dropout for regularization. We use standard binary cross-entropy loss with class balancing to handle the inherent imbalance in click data.

\subsubsection{Training Details}

We address class imbalance using weighted binary cross-entropy:
\begin{equation}
\mathcal{L} = -\frac{1}{N} \sum_{i=1}^{N} [w_+ y_i \log(\hat{y}_i) + w_- (1-y_i) \log(1-\hat{y}_i)]
\end{equation}
where $w_+ = N_-/N_+$ balances positive and negative samples.

Training parameters: Adam optimizer ($\alpha = 0.001$), batch size 2048, 10 epochs, dropout rate 0.2.

\subsubsection{Experimental Results}

Table~\ref{tab:behavior_results} shows performance improvements from behavior sequence integration.

The behavior-enhanced model achieves AUC of 0.6657 and log loss of 0.6012, representing meaningful improvements over the baseline approach. The dual embedding strategy effectively captures both category preferences and brand affinity patterns.

\subsubsection{Analysis}

Behavior sequences contribute most significantly for users with rich interaction histories. The fixed-length representation maintains computational efficiency while capturing essential temporal patterns. Parameter overhead remains modest at approximately 15

This architecture serves as an intermediate step toward more sophisticated sequence modeling approaches. The dual embedding strategy effectively captures both category preferences and brand affinity patterns, providing meaningful improvements over static feature models.

\subsubsection{Transformer-based CTR Model}

Building upon the success of behavior sequence integration, we developed a state-of-the-art transformer-based architecture that leverages multi-head attention mechanisms to capture complex temporal dependencies and interaction patterns in user behavior.

\textbf{Multi-Head Self-Attention}

The transformer architecture employs multi-head self-attention to model relationships between different behavior sequence elements:
\begin{align}
\text{Attention}(Q,K,V) &= \text{softmax}\left(\frac{QK^T}{\sqrt{d_k}}\right)V \\
\text{MultiHead}(Q,K,V) &= \text{Concat}(\text{head}_1, ..., \text{head}_h)W^O
\end{align}
where each attention head focuses on different aspects of the behavioral sequence.

\textbf{Enhanced Architecture Components}

The transformer model incorporates several advanced architectural features:

\emph{Positional Encoding}: We add learnable positional embeddings to preserve temporal order information:
\begin{equation}
\mathbf{H}_{pos} = \mathbf{H}_{behavior} + \mathbf{PE}
\end{equation}

\emph{Cross Network Integration}: Following DCN-V2 principles, we implement cross feature interactions:
\begin{equation}
\mathbf{x}_{l+1} = \mathbf{x}_0 \odot (\mathbf{W}_l \mathbf{x}_l + \mathbf{b}_l) + \mathbf{x}_l
\end{equation}

\emph{Advanced Sequence Aggregation}: Instead of simple mean pooling, we use masked weighted pooling to handle variable-length sequences:
\begin{equation}
\mathbf{s} = \frac{\sum_{i=1}^L \mathbf{H}_i \cdot m_i}{\sum_{i=1}^L m_i + \epsilon}
\end{equation}
where $m_i$ indicates whether position $i$ contains valid behavior data.

\textbf{Training Enhancements}

The transformer model employs several advanced training techniques:
- \emph{AdamW Optimizer}: Decoupled weight decay for improved generalization
- \emph{Cosine Annealing}: Warm restart scheduler for better convergence
- \emph{Gradient Clipping}: Prevents gradient explosion in deep networks
- \emph{Early Stopping}: Prevents overfitting with patience-based monitoring

\textbf{Experimental Configuration}

The final transformer configuration uses:
- Sequence length: 50 (extended from 20 for richer context)
- Embedding dimensions: 80 for categorical features, 48 for behavior sequences
- Multi-head attention: 8 heads across 4 transformer layers
- Deep MLP: [1024, 512, 256, 128] hidden units with GELU activation

\section{Results}

To assess baseline performance for CTR prediction, we compared three classical models: Logistic Regression, LightGBM, and a multi-layer perceptron (MLP). Evaluation metrics include AUC, Log Loss, and PR AUC, reflecting ranking accuracy, prediction calibration, and sensitivity to rare clicks.

Logistic Regression provided a simple and interpretable baseline but showed signs of underfitting due to its limited capacity to model nonlinear interactions. LightGBM, with its tree-based structure, improved performance by capturing complex feature relationships more effectively. MLP further outperformed both, leveraging embedding layers and deep architectures to model high-dimensional categorical features and achieve stronger overall results.

In summary, while each model offers trade-offs between complexity and performance, MLP delivered the best predictive accuracy among the three, and served as a solid foundation before we transitioned to more advanced behavior-sequence-based models in later stages.

Table~\ref{tab:sequence_model_results} presents the performance progression across our behavior-augmented models.

The transformer-based model achieves the highest performance with AUC of 0.6870, demonstrating the effectiveness of attention mechanisms in capturing complex behavioral patterns. The progressive improvement from static features (0.6442) to transformer architecture (0.6870) validates our sequential modeling approach.

\textbf{Computational Analysis}

The transformer model requires approximately 147M parameters, representing a significant increase from the basic behavior model. However, the attention mechanism provides interpretable insights into user behavior patterns, enabling better understanding of feature interactions.

\textbf{Model Insights}

Analysis of attention weights reveals that the model effectively identifies:
- Temporal patterns in user engagement (time-of-day effects)
- Sequential dependencies between product categories  
- Brand loyalty transitions over time
- Contextual relevance of historical interactions

These findings confirm that transformer architectures can effectively model the complex temporal dynamics inherent in user behavior for CTR prediction tasks.

\section{Discussion}

After evaluating the model using offline metrics, we designed a comprehensive A/B testing plan to verify the effectiveness of the constructed CTR prediction model in real-world applications. This experiment aims to evaluate whether an ad ranking strategy based on MLP model can significantly improve CTR performance in a real user environment, and to directly compare it with the existing system. Although this project is not yet capable of online deployment, the testing plan provides a clear execution framework and evaluation criteria for future staged rollouts.

\subsection{ A/B Test }

\subsubsection{Group Design}

The core objective of this A/B test is to evaluate whether our MLP-based CTR prediction model delivers real business value within the recommendation system. The specific goals include: Measure whether the MLP model outperforms the current ad recommendation mechanism in terms of CTR. Compare the impact of different recommendation strategies on user click behavior and provide data-driven decision support for system optimization and deployment.

We set the following experimental group structure:

\textbf{Group A (Control Group)}: Users will receive ad recommendations from the existing system, typically based on rule-based or random display logic.

\textbf{Group B (Experimental Group)}: Users will receive ads ranked based on the predictions of the MLP model, with the system displaying the Top-N ads with the highest predicted CTR.

\subsubsection{User Assignment Strategy}

To ensure good comparability between experimental groups and balanced data distribution across groups, we adopted a combined strategy of stratified random assignment and a cookie-based persistence mechanism for user allocation. Before the experiment begins, we will perform stratification based on key user profile variables such as gender, age group, geographic location, and device type. 

In addition, before and during the experiment, we will regularly perform sampling statistics on the experimental population to verify the overall balance of key variables—such as gender ratio, age distribution, geographic location, device type, and user activity level—across all groups. 

\subsubsection{Data Collection}

The collected data includes key behavioral fields such as user ID (user\_id), ad ID(adgroup\_id), click label(clk), and timestamp(log\_time). During the experiment, we will collect user behavior and impression data in real time to ensure accurate evaluation of key metrics, especially the CTR.In addition, we will collect certain user profile attributes such as gender, age, region, and device type for subsequent group balance validation and segmented analysis.

Data collection will be carried out through a combination of front-end tracking and back-end logging interfaces. All raw data will be centrally aggregated into a logging system or database, serving as the foundation for subsequent statistical analysis and model evaluation.  We will also conduct regular checks on data integrity and stability, focusing on issues such as missing data, continuity of metric recording, and abnormal fluctuations in click-through rate. 

\subsubsection{Evaluation Indicators and Statistical Analysis Methods} 
The main evaluation indicator of this experiment is the Click-Through Rate (CTR), which is defined as the ratio of the number of clicks to the number of exposures:
CTR = number of clicks / number of exposures.
We will count the average CTR of the control group and the experimental group and compare the difference between them.

To determine if the difference is statistically significant, we will use a two-sample z-test. The z-value, p-value, and 95\% confidence interval will be included in the report.

If there is sufficient data, we will also introduce additional metrics such as conversion rate (CVR), effective thousand exposure revenue (eCPM), page dwell time, bounce rate, etc., to more comprehensively assess user experience and business value.

\subsubsection{Estimation of sample size and experimental period} 
To ensure that the experiment has sufficient statistical power, we estimate the sample size based on the following parameters: the expected CTR of the control group is 3\%, the Minimum Detectable Effect (MDE) is 0.5\%, the significance level is alpha = 0.05, and the statistical power is 0.8. Use power.prop.test() to perform the sample size estimate. The estimation yields a minimum of about 17,000 users per group. If about 5,000 users can be included per day, the experiment is expected to last about 7 days.

\subsubsection{Experiment Follow-up and Go-live Strategy} 
After the experiment, we will decide whether to enter the grey-scale go-live process based on the following criteria: if the CTR of the experimental group is significantly better than that of the control group and has business significance, we will recommend the model strategy to go live; if the performance of multiple experimental groups improves, we will give priority to the solution with the largest CTR improvement; if no significant difference is observed, we will further analyze the importance of model features and assess whether there is any user behavioral bias. If no significant difference is observed, the importance of the model features will be further analyzed and evaluated to see if there is any user behavioral bias, data collection problems, or model underfitting.

\subsubsection{Risk Control and Open Issues} 
In the experiment planning process, we identified a number of potential risks and open issues, including whether the loss of user cookies or the replacement of devices will lead to cross-group errors, whether revenue-related indicators should be introduced in addition to CTR, and whether the experimental design should be adjusted accordingly in the event of fluctuations in user behavior caused by external factors such as holidays. These issues will be explored and discussed in subsequent iterations.

\subsection{Impact}

This project significantly improved the matching efficiency between users and ads by building an efficient click-through rate (CTR) prediction model. More accurate CTR predictions can help optimize ad sequencing and increase click-through rates, as well as increase ad revenue. At the same time, the improvement of model training and inference efficiency effectively reduces the consumption of computing resources, and enhances the sustainability and cost optimization capabilities of the commercial advertising system. In the model evaluation stage, we found that the MLP model performed the best in prediction performance, which provided a reliable reference and practical value for the deployment of CTR prediction models in future advertising systems.

In addition, CTR prediction and ad matching technology have also shown broad application prospects in the field of public health. Through this technology, health-related information can be disseminated more precisely to target populations, improving the effectiveness and coverage of public health interventions. The project further explores the issue of information equity in low-income groups and users in remote areas, and helps promote equity practices in public health. On the whole, this project embodies the organic combination between AI technology and social ethical goals, promotes the evolution of technology in the direction of "people-oriented", and expands the value space of CTR prediction model in public welfare scenarios.

\section{Conclusion}

Based on the Taobao advertising data set, this project systematically designed and evaluated a series of click rate (CTR) prediction models. Starting from the logical regression baseline, we gradually develop into a gradient-enhancing decision tree (LightGBM), a deep multi-layer sensor (MLP), and finally a behavior-enhanced Transformer architecture. The final transformer model integrates a multi-head self-attention mechanism to capture the time dependence in the user's behavior sequence. Its AUC reaches 0.687, and its logarithmic loss function (LogLoss) reaches 0.567. Compared with the baseline based on static characteristics, AUC increased by 6.64\%, highlighting the great value of integrated serial user signals. Although the absolute improvement of traditional indicators may seem bland, it is of extraordinary significance in large advertising platforms. Even a small AUC increase can bring significant revenue growth and measurable user participation.

In the process of this work, we have encountered some challenges related to technology and data. These problems include the sparseness of user behavior logs, the missing value of key features such as brands exceeding 30\%, the computing needs brought by the Transformer architecture with 147 million parameters, and the serious category imbalance, with an overall click rate of only 5.14\%. We meet these challenges through a combination of methods and strategies: using modal values for classification interpolation, category balance loss function, gradient cropping, and GPU acceleration training. In addition, we introduced a double-embedded layer to represent high-base characteristics and built cross-verified click-rate statistics to reduce data leakage.

In short, this project not only provides the most advanced click rate prediction model but also lays the foundation for scalable user behavior modeling methods in recommended systems across different application fields.

\section{Remaining Challenges and Future Directions}

\subsection{Challenges}
Although we have designed a complete set of A/B testing solutions to evaluate the effectiveness of MLP-based CTR prediction models in real-world scenarios, there are still many practical difficulties in implementing online A/B testing at this stage.

Firstly, we didn't have a production environment that could be deployed online. The real-time ad sequencing model needs to be deeply integrated with the advertising delivery system, and the current academic project environment cannot support real-time inference, log collection, and model version management, and cannot meet the basic technical requirements of online A/B testing.

Secondly, the scale of user traffic and data available is limited. Ensuring that A/B testing is statistically powerful requires a large number of active users and accurate traffic allocation, and we currently don't have access to commercial-grade ad networks to support data at that scale.

Thirdly, the team has limited resources and engineering capabilities. Building a complete online A/B testing process, including user triage, online logging, monitoring dashboards, and automated analysis, requires a professional engineering and O\&M team, which currently does not have the corresponding human resources support.

In addition to the limitations of A/B testing implementation, we also foresee challenges in a broader sense. However, in the actual online environment, user behavior patterns and ad inventory are constantly changing, and the performance of the model may drift, affecting the prediction accuracy. Secondly, the problem of data quality and bias also needs to be paid attention to, such as noise, missing values and sampling bias in user portrait and behavior log data, which may affect the feature learning effect in the model training process, as well as the fairness and reliability of the final evaluation results. Finally, in terms of interpretability and performance trade-off, although the MLP model performs best in offline evaluation, as a deep neural network, its model structure is relatively complex and the interpretability is insufficient, which is difficult to fully meet the requirements of business departments and regulators for model transparency and controllability, and it is necessary to further explore technical paths to improve the interpretability of the model in the future.

\subsection{Future Directions}

Based on our current work, we have proposed three strategic improvement paths. First of all, we will conduct a layered A/B test to empirically verify the actual effectiveness of the Transformer model. This experiment will compare our click-based ranking system with the existing platform logic and use strict statistical tests (such as z tests) to evaluate the improvement of core advertising indicators, including click-through rate (CTR), conversion rate (CVR) and cost per thousand impressions (eCPM). Secondly, we plan to optimize hyperparameters through automated machine learning (AutoML) frameworks such as AutoGluon, and combine the quantification of defect rates which is an important initiative of our Microsoft Ads partners to measure data anomalies (such as brand prices and missing values). Third, we will explore the cross-field adaptability of the framework, especially in personalized health communication systems. Participation in predictive targeted information (such as preventive care notifications) can expand the social influence of our methods.

\section{Application to Public Health}

Our CTR prediction models also hold significant potential value in the field of public health. Health-related advertisements can be accurately delivered to the people who need them most by combining CTR prediction with user-ad matching technology. For example, during flu season, the system can deliver vaccination advertisements to users in high-risk areas. For users who exhibit behavioral patterns related to psychological stress or sleep disturbances, targeted mental health support messages can be delivered. 

It is especially worth emphasizing the value of this technology for low-income populations and users in remote areas. This group of people usually lacks access to health information. We can use models to identify high-risk individuals and conduct precise interventions, and use the advertising system's high coverage and personalized push capabilities. This can improve the fairness and efficiency of public health interventions and promote a more equitable distribution of health resources in society.

\section{Acknowledgements}

Throughout the Capstone project, our six-person team has accumulated valuable experience under the guidance and support of multiple parties. We sincerely thank everyone who has helped us along the way.

First of all, we would like to sincerely thank our industry mentor, Dr.  Linxia Ren from Microsoft, for the professional insights and guidance she provided at every stage of the project. Her support helps us better understand the real business environment and continuously improve our model. We are especially grateful to her for taking the time to meet with us via Zoom and attend our Capstone summit in person.

We would also like to extend our felt thank to our faculty advisor, Dr. Hua Zhou, for his patient guidance and technical support. He generously spared time to attend all of our team meetings and provided valuable feedback and encouragement. We are especially grateful to him for helping us review and revise the presentation, which greatly improved the clarity and quality of our presentations.

Furthermore, we would also like to sincerely thank our teaching assistant, Bowen Zhang. He considerately assisted us in finding the data set, guided the implementation of the model, and answered our technical questions, enabling us to successfully complete each stage of the project.

Finally, we express our gratitude to all the instructors, staff, and supporters who have contributed to our Capstone project. Their guidance and encouragement are crucial to the successful completion of our project.

\newpage

\bibliographystyle{apalike}
\bibliography{example}

\appendix
\section{Appendix}

This appendix provides supplementary tables and figures referenced in the main report.

\subsection{figure}

\begin{figure}[htbp]
\centering

\subfloat[Top 10 Brands by Exposure]{
    \includegraphics[width=0.31\linewidth]{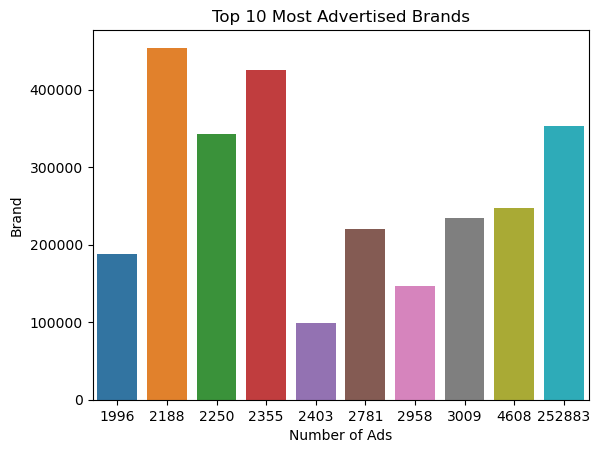}
    \label{fig:sub_top10brand}
}
\hfill
\subfloat[CTR by Price Level]{
    \includegraphics[width=0.31\linewidth]{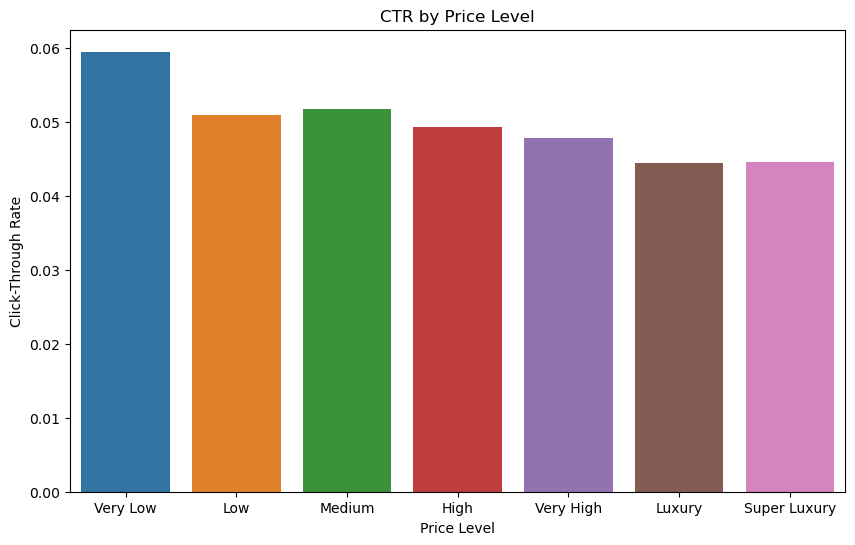}
    \label{fig:sub_price}
}
\hfill
\subfloat[Brand CTR Distribution]{
    \includegraphics[width=0.31\linewidth]{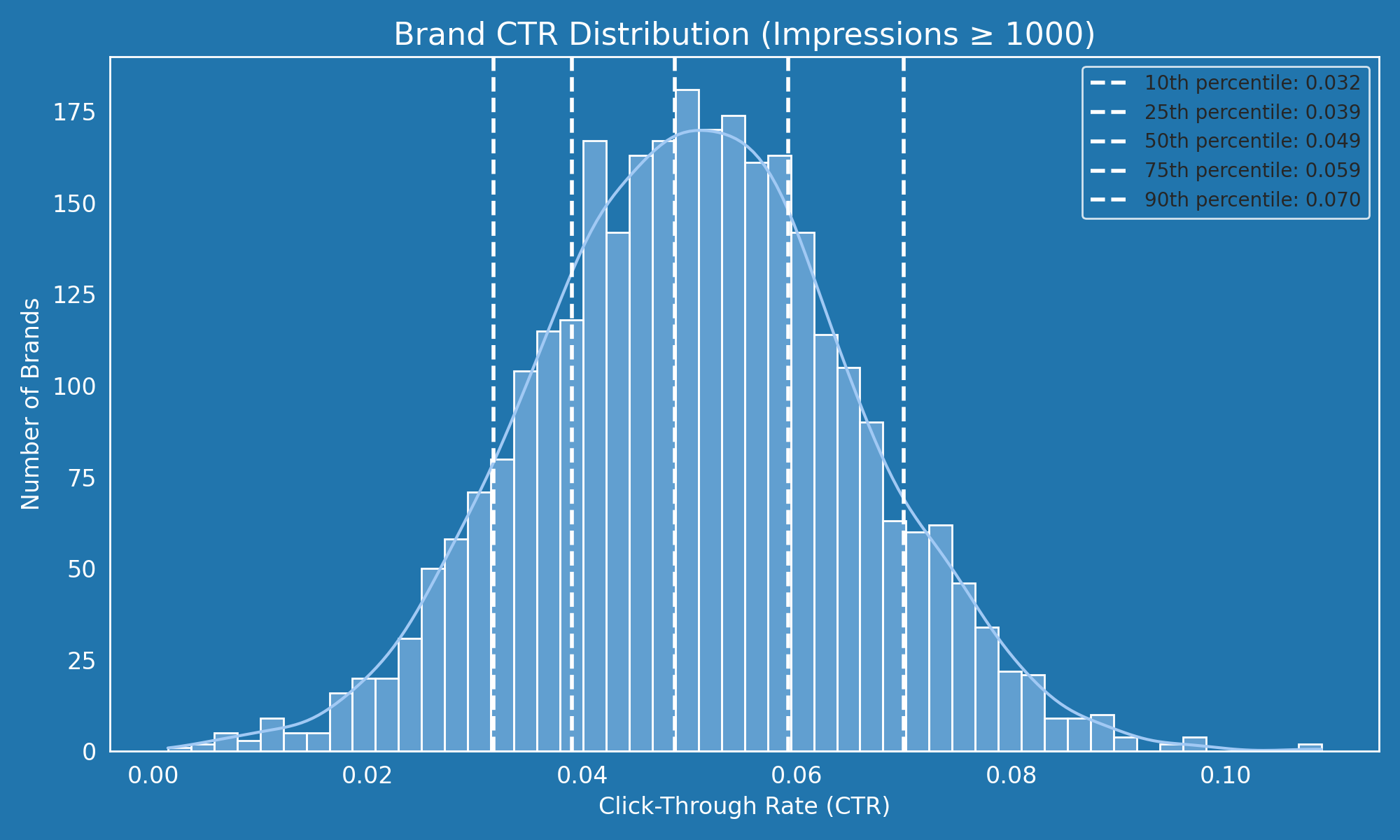}
    \label{fig:sub_brandctr}
}

\caption{Ad feature analysis: brand exposure, price gradient, and CTR distribution}
\label{fig:ad_feature}
\end{figure}

\subsection{table}

\begin{table}[h]
\centering
\caption{Performance comparison of behavior-enhanced model}
\label{tab:behavior_results}
\begin{tabular}{lcc}
\hline
Model & AUC & Log Loss \\
\hline
Baseline (feature-only) & 0.6442 & 0.6231 \\
+ Behavior sequences & 0.6657 & 0.6012 \\
\hline
Improvement & +3.34\% & -3.51\% \\
\hline
\end{tabular}
\end{table}

\begin{table}
  \caption{Model Comparison}
  \label{models-table}
  \centering
  \begin{tabular}{llll}
    \toprule
    Model Name     & AUC     & LogLoss  & PR AUC \\
    \midrule
    Logistic Regression & 0.659  & 0.655  &  0.097       \\
    LightGBM     & 0.660 & 0.194 &  0.099     \\
    Multilayer Perceptron     & 0.663  & 0.638      &  0.101 \\
    \bottomrule
  \end{tabular}
\end{table}

\begin{table}[h]
\centering
\caption{Performance comparison of sequence-enhanced models}
\label{tab:sequence_model_results}
\begin{tabular}{lcc}
\hline
Model Architecture & AUC & Log Loss \\
\hline
Baseline MLP (static features) & 0.6442 & 0.6231 \\
+ Behavior sequences (MLP) & 0.6657 & 0.6012 \\
+ Transformer attention & 0.6870 & 0.5674 \\
\hline
Total improvement & +6.64\% & -8.94\% \\
\hline
\end{tabular}
\end{table}


\end{document}